\documentclass{article}

\usepackage[nonatbib]{neurips_2024}
\usepackage{neurips_2024}

\usepackage[utf8]{inputenc} 
\usepackage[T1]{fontenc}    
\usepackage{hyperref}
\hypersetup{colorlinks,allcolors=black}
\usepackage{booktabs}       
\usepackage{amsfonts}       
\usepackage{nicefrac}       
\usepackage{microtype}      
\usepackage{xcolor}         
\usepackage{graphicx}
\usepackage{booktabs}
\usepackage{caption}
\usepackage{lipsum} 
\usepackage{wrapfig}
\usepackage{amssymb }
\usepackage{multirow}
\usepackage{url}            
\usepackage{booktabs}       
\usepackage{amsfonts}       
\usepackage{amsmath}
\usepackage{amsthm}
\usepackage{amssymb}
\usepackage{nicefrac}       
\usepackage{microtype}      
\usepackage{xcolor}         
\usepackage[font=small,labelfont=bf]{caption}
\usepackage{enumitem}
\usepackage{wrapfig}
\usepackage{listings}
\usepackage{caption}
\usepackage{mathtools}
\usepackage{xcolor}
\usepackage{graphicx} 
\usepackage{amsmath} 
\usepackage{amsfonts}
\usepackage{mathtools}
\usepackage{bbm}
\usepackage{algorithm}
\usepackage{algpseudocode}
\usepackage{cleveref}
\usepackage{subcaption}
\usepackage{wrapfig}
\usepackage{ amssymb }
\usepackage{multirow}
\usepackage[normalem]{ulem}
\usepackage{xspace}
\usepackage{float}
\usepackage{tabularx}
\usepackage[normalem]{ulem}
\usepackage{fontawesome}
\useunder{\uline}{\ul}{}

\newcommand{\Envelope}{\faEnvelopeO}

\newcommand{\Wup}{W_{\text{up}}}

\newcommand{\Wdown}{W_{\text{down}}}
\newcommand{\Wgate}{W_{\text{gate}}}

\begin{document}

\title{Turbo Sparse: Achieving LLM SOTA Performance with Minimal Activated Parameters}

\author{\rm
\begin{tabular}{c}
\normalsize Yixin Song$^{\text{1}}$, Haotong Xie$^{\text{1}}$, Zhengyan Zhang$^{\text{2}}$, Bo Wen$^{\text{1}}$, Li Ma$^{\text{3}}$, Zeyu Mi$^{\text{1}}$\Envelope, and Haibo Chen$^{\text{1}}$ \\[5pt]
\normalsize $^{\text{1}}$Institute of Parallel and Distributed Systems (IPADS), Shanghai Jiao Tong University \\
\normalsize $^{\text{2}}$Department of Computer Science and Technology, Tsinghua University \\
\normalsize $^{\text{3}}$Shanghai Artificial Intelligence Laboratory
\end{tabular}
}

\maketitle

\renewcommand*{\thefootnote}{{\Envelope}}
\footnotetext[1]{ Corresponding author: Zeyu Mi (\url{yzmizeyu@sjtu.edu.cn}).}
\renewcommand{\thefootnote}{\arabic{footnote}}
\setcounter{footnote}{0}

\begin{abstract}
Exploiting activation sparsity is a promising approach to significantly accelerating the inference process of large language models (LLMs) without compromising performance. However, activation sparsity is determined by activation functions, and commonly used ones like SwiGLU and GeGLU exhibit limited sparsity. Simply replacing these functions with ReLU fails to achieve sufficient sparsity. Moreover, inadequate training data can further increase the risk of performance degradation.
To address these challenges, we propose a novel dReLU function, which is designed to improve LLM activation sparsity, along with a high-quality training data mixture ratio to facilitate effective sparsification. Additionally, we leverage sparse activation patterns within the Feed-Forward Network (FFN) experts of Mixture-of-Experts (MoE) models to further boost efficiency.
By applying our neuron sparsification method to the Mistral and Mixtral models, only 2.5 billion and 4.3 billion parameters are activated per inference iteration, respectively, while achieving even more powerful model performance. Evaluation results demonstrate that this sparsity achieves a 2-5× decoding speedup. Remarkably, on mobile phones, our TurboSparse-Mixtral-47B achieves an inference speed of 11 tokens per second. Our models are available at \url{https://huggingface.co/PowerInfer}.

\begin{figure}[!ht]
\centering
\includegraphics[width=0.5\linewidth]{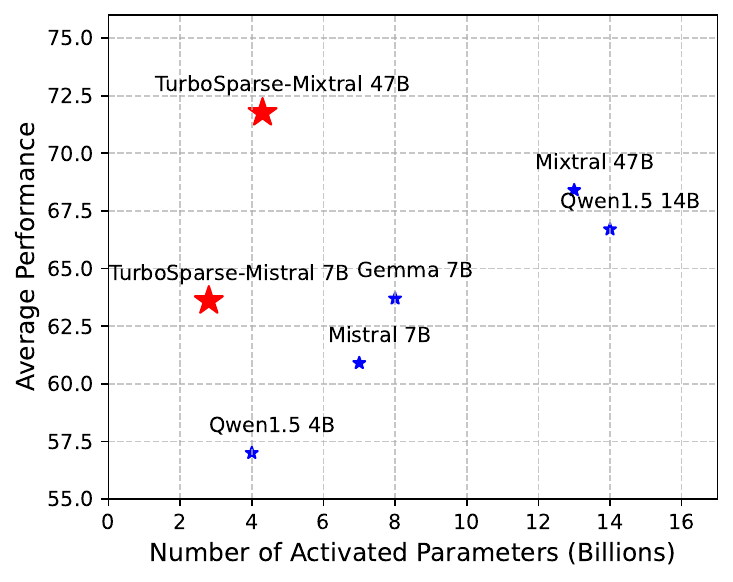}
\caption{
Comparison on the Open LLM Leaderboard shows that our dReLU-based sparsified models, particularly TurboSparse-Mixtral-47B, consistently outperform similar models. 
}
\label{fig:takeaway}
\end{figure}
\end{abstract}
\section{Introduction}

Large Language Models (LLMs) have achieved remarkable results, demonstrating emergent natural language abilities as the number of model parameters scales~\cite{brown2020language, zhang2022opt}. These models have pushed the state-of-the-art performance across a wide range of downstream applications, such as QA and coding.
However, most LLMs, such as Llama~\cite{touvron2023llama}, Mistral~\cite{jiang2023mistral}, and Gemma~\cite{team2024gemma}, utilize all of their parameters during inference. These are known as dense models. The escalating demand for computational resources by dense models has become a significant barrier to the development of powerful and accessible AI, given the substantial costs involved.

To address the efficiency issues inherent in dense models, conditional computation~\cite{bengio2013deep,bengio2015conditional} has emerged as a crucial approach, which refers to activating part of the neurons in a network. There are two primary methods to achieve conditional computation. Mixture-of-Experts (MoE)~\cite{fedus2022switch, lepikhin2020gshard} is the first promising method, which introduces conditional computation by manually setting constraints on the model architecture prior to training, such as determining the number of experts to activate. This technique selectively activates specific parts of the model in response to particular inputs through a process known as expert routing, resulting in significant efficiency improvements. For instance, Switch Transformer~\cite{fedus2022switch} has scaled the model to the trillion-parameter level without increasing computational FLOPs significantly.
Another promising method is utilizing the natural emergence of sparse activation due to the ReLU activation function~\cite{li2022lazy}, which naturally outputs zero elements that have no contribution in computation results. This activation sparsity presents a significant opportunity for efficient inference.
Deja Vu~\cite{dejavu} utilizes that sparsity exists in dense models due to ReLU to achieve 2$\times$ speedups. PowerInfer~\cite{song2023powerinfer} achieving up to 11$\times$ speedups for deploying larger LLMs in a single consumer-grade GPU setting.

Recent LLMs typically prefer activation functions such as GELU~\cite{hendrycks2016gaussian} and Swish~\cite{ramachandran2017searching}. However, these functions do not significantly promote activation sparsity and are challenging to accelerate with conditional computation. To address this, ReLUfication~\cite{mirzadeh2023relu}, an existing state-of-the-art method, replaces the original activation function with ReLU and continues with pretraining. Despite its potential, this approach often struggles to achieve the desired levels of activation sparsity and may risk performance degradation~\cite{lee2024cats,sparsellm}.

We argue that the failure of existing ReLUfication methods can be attributed to two main reasons. First, simply substituting SwiGLU with ReGLU is inefficient, as it only increases sparsity from 40\% to around 70\%. It suggests that a deeper investigation into the model architecture is necessary to achieve higher levels of sparsity. Second, the limited diversity of pretraining data and the insufficient number of training tokens in current approaches lead to incomplete capability recovery~\cite{mirzadeh2023relu,lee2024cats}. As a result, expanding the diversity of pretraining datasets and increasing the number of training tokens are critical steps towards enhancing model performance.

To address these challenges, we first conduct a comprehensive analysis of the existing ReLUfication approach and identify that its shortcomings stem from the negative activations in the GLU component. Therefore, we propose an efficient activation function named dReLU. We apply dReLU in the pretraining of small-scale LLMs, alongside SwiGLU, and our findings indicate that LLMs using dReLU match the performance of those using SwiGLU, while also achieving close to 90\% sparsity. Additionally, we collect a diverse range of pretraining corpora from the open-source community, including web, code, and mathematical datasets, to enhance the effectiveness of ReLUfication.

Meanwhile, we also conduct a sparsity analysis on MoE-based LLMs. Interestingly, we observe that the feed-forward networks (FFNs) within the experts remain sparsely activated, similar to the behavior exhibited by dense LLMs. This phenomenon suggests an opportunity to further accelerate inference speed by combining MoE techniques with ReLU-based sparse activation.

\begin{figure}[!ht]
\centering
\includegraphics[width=0.7\linewidth]{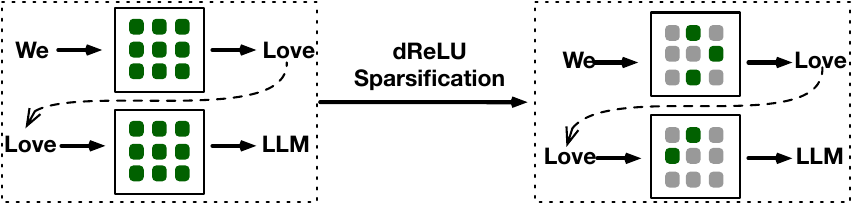}
\caption{
Example of dReLU sparsification. The left figure illustrates the original dense activation where every input activates all neurons, while the right is our sparsified LLMs, where each input activates only a small subset of neurons.
}
\label{fig:overview}
\end{figure}

To validate the effectiveness of our proposed method, we implemented it on the Mistral-7B and Mixtral-47B models, converting them to TurboSparse-Mistral-47B and TurboSparse-Mixtral-47B, respectively. Extensive experiments across a wide range of downstream tasks demonstrate (Figure ~\ref{fig:takeaway}) that our enhanced models not only meet but often surpass the performance of their original counterparts.

Remarkably, in the TurboSparse-Mistral-7B model, we increase the average sparsity of the FFN to 90\% while enhancing model capabilities. In MoE models, we further improve the sparsity in the TurboSparse-Mixtral-47B, originally introduced due to expert routing, from 75\% to 97\% by incorporating sparse neuron activations. This substantial increase in sparsity significantly reduces FLOPs during the inference process.

Finally, we integrate our two new models with PowerInfer to evaluate the inference speed. Performance evaluation reveals that our models deliver an average 2.83$\times$ generation speedup.

The key contributions of this paper include:
\begin{itemize}
    \item \textbf{Efficient dReLU activation function:} Our method utilizes fewer than 150B tokens, representing less than 1\% of the typical pretraining tokens (commonly 15T tokens~\cite{cai2024internlm2}).
    \item \textbf{Sparse activated models:} We will release our sparsely-activated TurboSparse-Mistral-7B and TurboSparse-Mixtral-47B models. Both models demonstrate better performance compared to the original versions.
    \item \textbf{Practical inference speedup:} Evaluation shows that with our models, we can achieve a 2-5$\times$ speedup. Notably, we can achieve up to 10 tokens/s even without a GPU on TurboSparse-Mixtral-47B.
\end{itemize}

\section{Related Work and Background}
\paragraph{Efficient Inference of LLMs.}
Efficient LLM inference poses challenges that necessitate a synergistic combination of algorithmic and systemic approaches.
From an algorithmic standpoint, researchers have explored various methods to reduce computation and memory overheads, including compressing models~\cite{DBLP:conf/nips/MichelLN19,DBLP:conf/nips/YaoAZWLH22,GPTQ,AWQ,DBLP:conf/icml/XiaoLSWDH23}, modifying model structures~\cite{GQA,DBLP:journals/corr/abs-2312-00752}, and speculative decoding methods~\cite{DBLP:conf/icml/LeviathanKM23,DBLP:journals/corr/abs-2302-01318, cai2024medusa}.
On the systemic front, there are efforts that effectively integrate the features of downstream hardware and upper-level models to maximize the efficiency of computation and memory utilization~\cite{DBLP:conf/sc/AminabadiRALLZRSZRH22,DBLP:conf/icml/RajbhandariLYZA22,DBLP:conf/ppopp/FangYZZ21,DBLP:conf/osdi/YuJKKC22}, leading to the development of more efficient frameworks like vLLM~\cite{vLLM}.

Sparse activation, in particular, has emerged as a research area that demands an even tighter integration of algorithmic and systemic approaches. The selection of activation functions and the construction of activation predictors are algorithmic problems, while fully exploiting the sparse activation of LLMs on specific hardware is a systemic challenge. By leveraging sparse activation, researchers have achieved promising results in building efficient LLM inference systems~\cite{dejavu,song2023powerinfer}.

\paragraph{Mixture-of-Experts (MoE).} MoE techniques induce effective sparsity in LLMs by determining which subset of subnetworks (referred to as "experts") to activate during the inference pass, often through a trained "router" subnetwork. This approach allows the model to enhance its capacity without escalating the computational expenses~\cite{lepikhin2020gshard, shazeer2017outrageously}.

\paragraph{Intrinsic Activation Sparsity.} Intrinsic activation sparsity is known to be present in LLMs that utilize ReLU family nonlinearities in their MLP blocks~\cite{zhang2021moefication, li2022lazy}. This phenomenon has been explored to accelerate inference speed and reduce memory usage~\cite{song2023powerinfer, dejavu, liu2024ffsplit}. With this phenomenon, each neuron can be viewed as an expert to reduce the computation overhead.

\paragraph{Gated-MLP Blocks.} We now delve into the components of LLMs that our study aims to analyze: the Gated-MLP blocks, which are commonly used. A Gated-MLP block consists of three fully connected layers and performs the following computation:
\begin{equation}
\label{eqn:gated-MLP}
\begin{aligned}
\text{Gate}(x) &\coloneqq F_{act} (x \Wgate) \\
\text{Up}(x) &\coloneqq x \Wup \\
\text{Combined}(x) &\coloneqq \text{Gate}(x) * \text{Up}(x) \\
\text{Gated-MLP}(x) &\coloneqq \text{Combined}(x) \Wdown
\end{aligned}
\end{equation}

where $F_{act}$ represents different activation functions like ReLU~\cite{agarap2019deep}, SiLU~\cite{loshchilov2017decoupled}.

\section{Analysis}
\label{Analysis:drelu}

\subsection{Limitations about Existing ReLUfication}
\begin{figure}[!ht]
\centering
\includegraphics[width=1\linewidth]{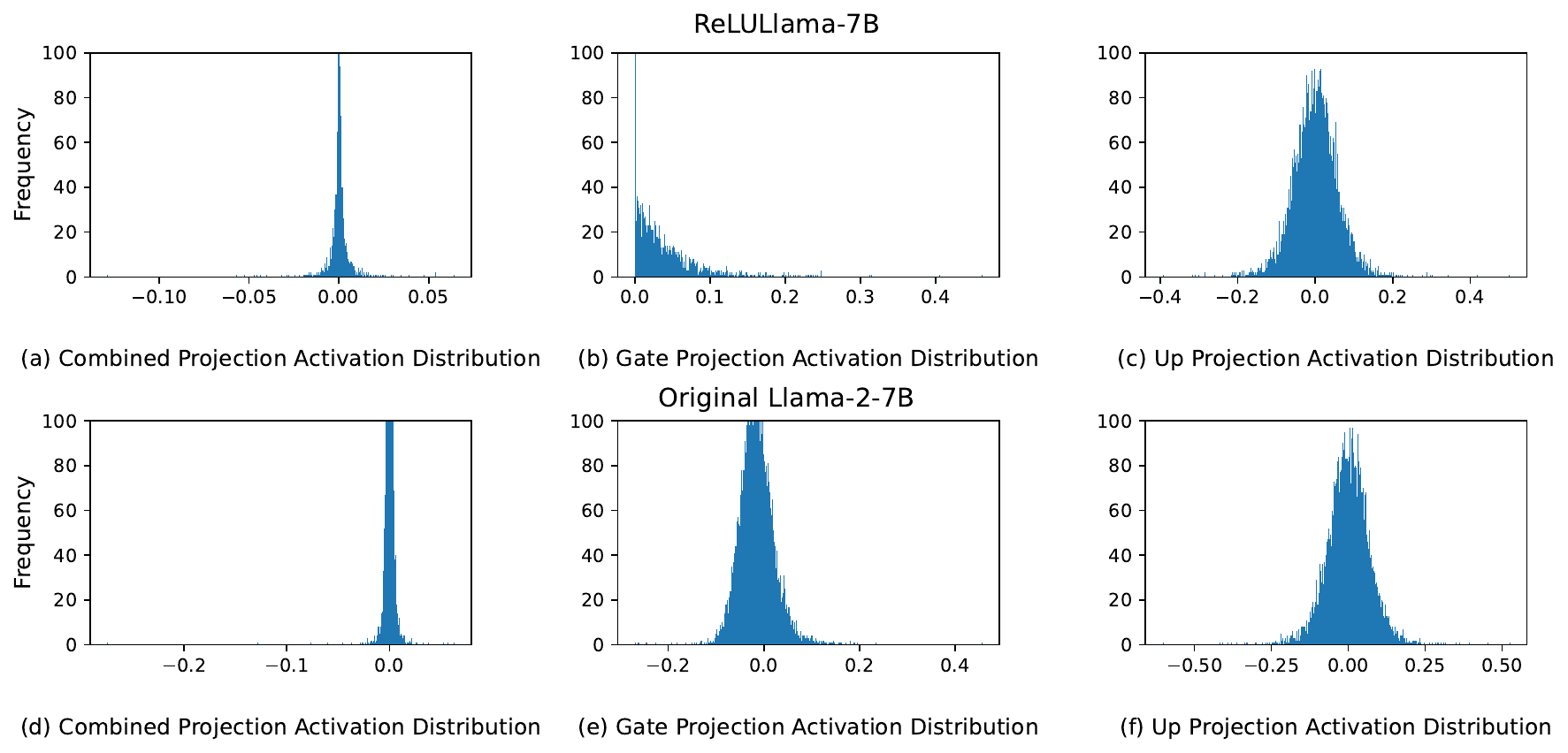}
\caption{
Post-activation distribution of ReLULlama and Llama-2-7B in layer 0.
}
\label{fig:activation}
\end{figure}
\begin{wraptable}{r}{6.3cm}
\centering
\vspace{-25pt}
\caption{Model Sparisty compared ReLULlama with Llama-2-7B}
\label{tbl:sparsity}
\begin{tabular}{lcr}\toprule
Model & Sparisty \\\midrule
Llama-2-7B & 40\%  \\
ReLULlama-7B & 67\% \\
ShiftedReLULlama-7B & 71\%  \\\bottomrule
\end{tabular}
\vspace{-10pt}
\end{wraptable}
We first evaluate the sparsity of ReLULlama-7B~\cite{sparsellm} and the original Llama-2-7B~\cite{touvron2023llama}, as shown in Table~\ref{tbl:sparsity}. The results reveal that existing ReLUfication methods can only improve the sparsity from 40\% to 67\%, indicating their limited effectiveness in significantly enhancing model sparsity.

To investigate the underlying reasons for this limitation, we profile the activation distribution of the gate and up projection components separately in ReLULlama-7B and Llama-2-7B, as illustrated in Figure~\ref{fig:activation}. The figure shows that after ReLUfication, the combined activation becomes more concentrated around 0, with the sparsity increasing to 67\%. This can be attributed to the ReLU activation function applied after the gate weight, which masks all negative activations to zero.

To further push the sparsity, shifted-ReLU~\cite{mirzadeh2023relu} has been proposed, which adjusts the threshold of ReLU function to mask out more activations in the gate projection. However, the improvements brought by this method are limited. Another line of work is to adopt progressive sparsity regularization to the intermediate output to introduce more zero activation output~\cite{song2024prosparse}. However, this method carries the risk of performance degradation.

Existing ReLUfication methods primarily focus on modifying the gate component. Different from previous work, we find that existing ReLUfication doesn't alter the activation distribution of the up projection component, as shown in Figure~\ref{fig:activation}(c) and (f).
According to the definition of Gated-MLP (Equation \ref{eqn:gated-MLP}), the gate and up projection components jointly influence the sparsity of neuron activations in parallel. However, a significant number of activation values in the up projection component remain less than 0. This suggests that masking the outputs of the up and gate matrices that are less than 0 as inactive could introduce stronger sparsity without sacrificing non-linear capabilities. This observation motivates us to explore the possibility of further enhancing model sparsity by modifying the up projection.

\subsection{dReLU}

We introduce a new activation function, named dReLU (Equation \ref{eqn:drelu}), where ReLU is applied after both the up- and gate-projection\footnote{We omit the bias in both the up- and gate-projection to match the form of Equation \ref{eqn:gated-MLP}.}.
\begin{equation}
\label{eqn:drelu}
\begin{aligned}
    \text{Combined}_{\text{dReLU}}(x) & \coloneqq \max(0, x\Wgate) * \max(0, x\Wup)
\end{aligned}
\end{equation}

\begin{figure}[h]
    \begin{minipage}{0.5\textwidth}
        \centering
        \includegraphics[width=\textwidth]{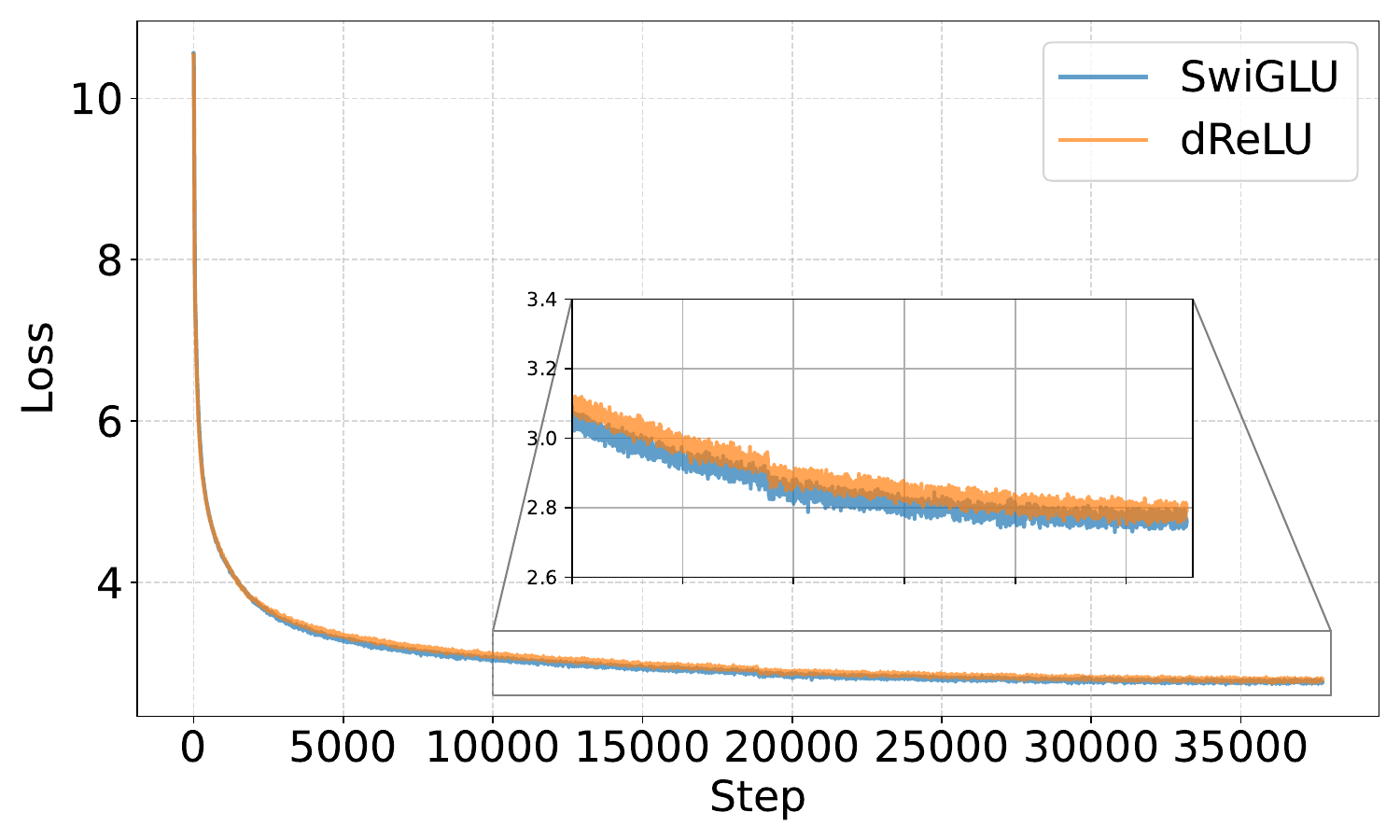}
        \caption{Training loss of small models with different activation functions.}
        \label{fig:loss_curve}
    \end{minipage}%
    \begin{minipage}{0.5\textwidth}
        \centering
        \begin{table}[H]
            \centering
            \begin{tabular}{lcc}
                \toprule
                Activation & Training Loss & Validation PPL \\ \midrule
                dReLU & 3.154 & 28.45 \\
                SwiGLU & 3.146 & 28.48 \\ \bottomrule
            \end{tabular}
            \caption{Validation and training loss on different activations.}
            \label{tab:50M_average}
        \end{table}
    \end{minipage}
\end{figure}

To demonstrate the effectiveness and performance of dReLU, we conducted an experiment comparing 300M-parameter decoder-only architecture models using dReLU and SwiGLU, both pretrained under the fineweb dataset~\cite{penedo2024fineweb} for 5B tokens. Refer to Appendix \ref{sc:300} for the detailed model architecture hyperparameters.  The evaluation result is shown in Table \ref{tab:50M_average}.

Our findings reveal models employing the dReLU structure exhibit similar convergence compared to those using the SwiGLU structure. Notably, we evaluate the perplexity of both models on Wikitext-2~\cite{merity2016pointer}. DReLU-based models show slightly better performance on WikiText-2~\cite{merity2016pointer}. 

Figure~\ref{fig:loss_curve} illustrates the loss curves during training, demonstrating that models with the dReLU activation function achieve similar convergence ability compared to their SwiGLU counterparts. To further validate this observation, we evaluate the perplexity of these models on the Wikitext2 dataset. As shown in Table~\ref{tab:50M_average}. Notably, although SwiGLU-based model has lower training loss, dReLU based model has lower validation perplexity. These results provide strong evidence that adopting the dReLU structure does not compromise model performance. We evaluate on more downstream tasks in Appendix \ref{sc:300}.

Another question we need to address is the dReLU-based model's sparsity. To investigate the sparsity of the dReLU-based model, we propose a methodology for measuring and evaluating a model's performance under different sparsity levels. Our approach involves selecting the top-$k$\% of values activated by dReLU or other activation functions based on their absolute magnitude, as described in Equations~\ref{eqn:mask} and \ref{eqn:sparsify-gated-MLP}.
\begin{equation}
\begin{aligned}
\label{eqn:mask}
\text{Mask}(x) \coloneqq \text{Top}_k(|\text{Combined}(x)|)
\end{aligned}
\end{equation}
\begin{equation}
\begin{aligned}
\label{eqn:sparsify-gated-MLP}
\text{Gated-MLP}(x) \coloneqq (\text{Combined}(x) * \text{Mask}(x)) \Wdown
\end{aligned}
\end{equation}
where $\Wdown$ represents the down-projection matrix. By varying the value of $k$, we can control the sparsity level of the model. To assess the impact of sparsity on model performance, we evaluate the dReLU-based model on a range of downstream tasks at different sparsity levels. This allows us to determine the optimal sparsity-performance trade-off for the dReLU activation function.
Table~\ref{tab:ppl_sparsity} presents the perplexity of the dReLU-based and SwiGLU-based models on WikiText-2 across various sparsity levels. The results demonstrate that the dReLU activation function enables high sparsity without significant degradation in performance, maintaining competitive performance even at 90\% sparsity. In contrast, the SwiGLU-based model experiences a severe increase in perplexity as sparsity reaches 80\%.

\begin{table}[t]
\centering
\begin{tabular}{lccccc}
\toprule
\textbf{Top-k\%} & \textbf{0} & \textbf{50\%} & \textbf{80\%} & \textbf{85\%} & \textbf{90\%} \\ \midrule
dReLU & 28.45 & 28.45 & 28.45 & 28.65 & 29.19 \\
SwiGLU & 28.48 & 28.62 & 36.28 & 48.55 & 112.36\\ \bottomrule
\end{tabular}
\caption{WikiText-2 perplexity on different sparsity on different models.}
\label{tab:ppl_sparsity}
\end{table}
\section{Are Neurons in Expert still Sparsely Activated?}
Previous work has shown that dense LLMs with different activation functions (ReLU, SwiGLU, etc.) exhibit the property of sparse activation~\cite{zhang2024relu,dejavu, lee2024cats}. However, the analysis is limited to dense models. Despite the intuitive assumption that partitioning FFNs into different experts within an MoE model would result in denser activations within each expert, it remains unclear whether this sparsity phenomenon persists in MoE models. In this section, we select representative MoE models and commonly used downstream tasks to investigate whether this sparsity phenomenon still exists in MoE models. We utilize the same method in \ref{Analysis:drelu} to control the sparsity in each expert. 

\paragraph{Models.} We select Deepseek-MoE~\cite{dai2024deepseekmoe}, Qwen1.5-MoE~\cite{qwen} and Mixtral~\cite{jiang2024mixtral} as the models for our experiments. We also add Llama-2-7B as for comparison.

\begin{figure}[ht]
    \vspace{-1mm}
    \subfloat[] {
    \includegraphics[width=0.45\linewidth]{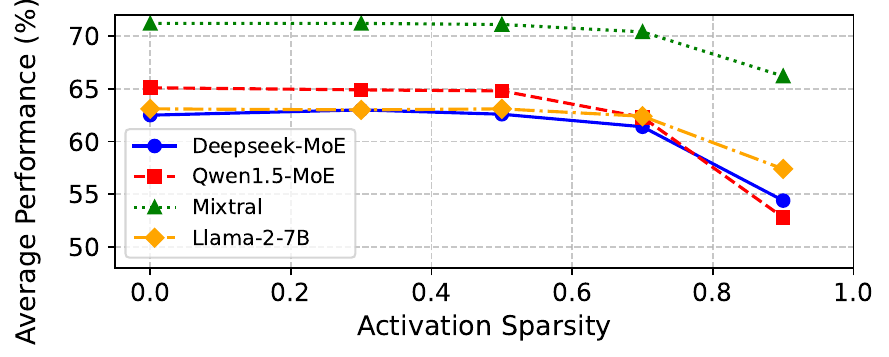}
    }
    \subfloat[] {
        \includegraphics[width=0.48\linewidth]{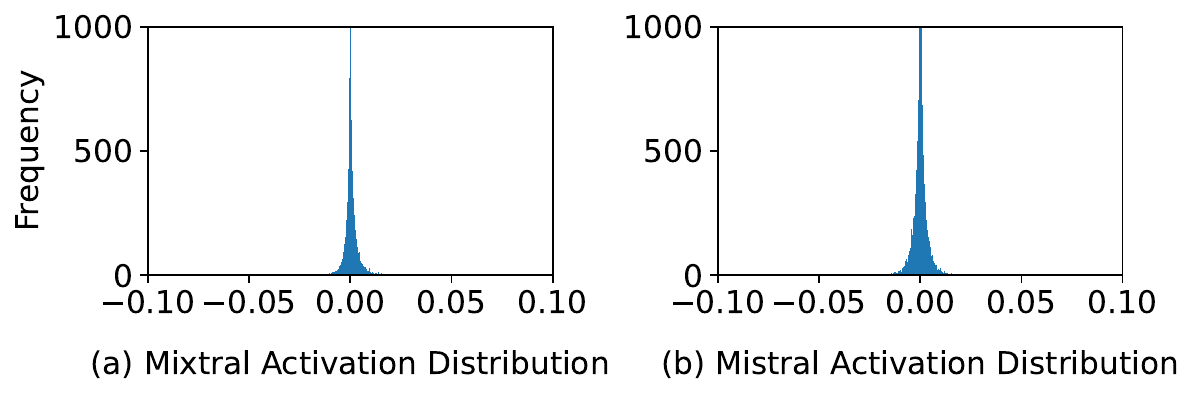}
    }
    \caption{\small{(a) Performance of MoE models with regard to activation sparsity. The impact of activation sparsity on the performance is negligible until the sparsity ratio is larger than 0.5. (b) Activation distribution of Mixtral and Mistral.}}
    \label{fig:moe}
\end{figure}

We first study the performance with regard to the sparsity ratio, as shown in Figure \ref{fig:moe} (a)\footnote{Each performance score is reported as an average of Winogrande~\cite{winogrande}, TruthfulQA~\cite{lin2021truthfulqa}, PIQA~\cite{Bisk2020}, LAMBADA~\cite{lambada}, ARC-Easy~\cite{arc}, and ARC-Challenge~\cite{arc}.}. Specifically, the performance only drops by about 1\%-2\% when the sparsity ratio is 0.5. This trend suggests that MoE models exhibit similar sparsity compared to dense models.

Further, we profile the activation patterns of Mistral and Mixtral, a pair of popular dense LLM and MoE LLM, as shown in Figure \ref{fig:moe} (b). We find that both LLMs show a similar pattern where activations are concentrated around 0, which is consistent with previous analysis of dense LLMs. The sparsity in experts also implies that every neuron in the same expert has different functionality. This finding applies to all layers and experts, as detailed in Appendix \ref{pe:MoE}. We report this interesting observation and leave further analysis for future work.

Inspired by our discoveries in MoE models, we are convinced that ReLUfication can be extended to MoE models and is not restricted to dense models. As the proportion of FFN weights in MoE models increases, the FLOP reduction achieved through ReLUfication will be even more pronounced.




\section{dReLU Sparsification}
In the previous section, we have demonstrated that dReLU can be a better choice for ReLUfication. 
The  main question now is whether dReLU based ReLUfication can recover the original model's performance while achieving higher sparsity. The following sections will discuss the experiments that aimed at answering this question.

\paragraph{Experimental setup.} We consider two representative models: Mistral-7B and Mixtral-47B. We substitute the original SwiGLU based FFN with dReLU based FFN and then continue pretraining.  

\paragraph{Pretraining datasets.} Due to the ReLUfication process, the restoration of model capability is closely related to the corpus used for recovery training. We collected as much corpus as possible from the open-source community for training, such as Wanjuan-CC~\cite{qiu2024wanjuancc}, open-web-math~\cite{paster2023openwebmath}, peS2o~\cite{peS2o}, Pile~\cite{gao2020pile}, The Stack~\cite{Kocetkov2022TheStack}, GitHub Code~\cite{codeparrot-github-code} and so on. The detailed mixture ratio is as shown in the following table~\ref{tab:data_mixture}:
\begin{table}[ht]
\centering
\begin{tabular}{llr}
\toprule
\textbf{Category} & \textbf{Dataset} & \textbf{Percentage} \\
\midrule
\multirow{4}{*}{Academic } & Pile-Arxiv & \multirow{3}{*}{2.69\%} \\
& Pile-PubMed & \\
& Pile-Philpapers & \\
& Dolma\_peS2o & 7.83\% \\
\midrule
\multirow{3}{*}{Web} & Wanjuan-CC & 73.55\% \\
& RedPajama-Wiki & 0.65\% \\
& Pile-OpenWebtext2 & 0.44\% \\
\midrule
\multirow{2}{*}{Books} & RedPajama-books & 6.54\% \\
& Pile-PG19 & 0.37\% \\
\midrule
\multirow{2}{*}{Math } & Open-web-math & 1.46\% \\
& Proof-pile-2 & 0.76\% \\
& algebraic-stack & 0.54\%\\
\midrule
\multirow{2}{*}{Code } & Starcoder-Java & 0.77\%\\
& Starcoder-C\# & 0.74\%\\
& Starcoder-Typescript & 0.52\%\\
& Starcoder-remaining & 1.73\% \\
& GitHub-Code & 1.41\%\\
\bottomrule
\end{tabular}
\caption{Detailed data mixture}
\label{tab:data_mixture}
\end{table}

\paragraph{SFT datasets.} After pretraining, we utilize the high-quality SFT datasets to further improve our model's performance, including orca-math-word-problems~\cite{mitra2024orcamath}, bagel~\cite{bagel}.

\paragraph{Hyper-parameters.} The hyperparameters for our ReLUfication are based on empirical results from previous works~\cite{zhang2024relu}. We utilize the llm-foundry framework for training~\cite{llmfoundry} and employ FSDP parallelism.

Our models are trained using the AdamW optimizer~\cite{loshchilov2017decoupled} with the following hyper-parameters: $\beta_1 = 0.9$ and $\beta_2 = 0.95$. We adopt a cosine learning rate schedule and use the default values for weight decay and gradient clipping (see Table~\ref{tab:hyperparameters_small} for more details). In total, we pretrain our models on 150B tokens.
\begin{table}[htbp]
    \centering
    \normalsize
    \caption{ 
      Details of training hyper-parameters.
    }
    \scalebox{1.0}{
    \begin{tabular}{lcccccccc}
      \toprule
      \textbf{Sequence Length}  & \textbf{Batch Size}& \textbf{Learning Rate} & \textbf{Warmup Steps} & \textbf{Hardware}\\
  
      \midrule    
      4,096       & 2,048       & $5e^{-5}\rightarrow5e^{-6}$    & 1000  & 64 A800-80G GPUs  \\
      \bottomrule
    \end{tabular}
    }
    \label{tab:hyperparameters_small}
\end{table}
\section{Experiments Results}
\subsection{Downstream Tasks Performance}

We measure our sparsified models' performance on tasks included in OpenLLM Leaderboard which include 25-shot Arc-Challenge~\cite{arc}, 10-shot Hellaswag~\cite{zellers2019hellaswag}, 5-shot MMLU~\cite{hendryckstest2021}, 0-shot TruthfulQA~\cite{lin2021truthfulqa}, 5-shot Winogrande~\cite{winogrande} and 8-shot GSM8K~\cite{cobbe2021gsm8k}. In addition, we also follow Llama 2's evaluation task included commonsense reasoning tasks. We report the average of PIQA~\cite{Bisk2020}, SCIQ~\cite{SciQ}, ARC easy~\cite{arc}, OpenBookQA~\cite{OpenBookQA2018}. We compare our models to several external open-source LLMs, including Gemma-2B~\cite{team2024gemma}, Mistral-7B~\cite{jiang2023mistral} and Mixtral-47B~\cite{jiang2024mixtral}.
\begin{table}[ht!]
\centering
\begin{tabular}{ccccccc}
\toprule
& Gemma & Mistral & TurboSparse & Mixtral & TurboSparse  \\
& -2B & -7B & -Mistral-7B & -47B & -Mixtral-47B \\
\midrule
\# Total Params & 2B & 7B & 7B & 47B & 47B\\
\# Activate Params & 2B & 7B & 2.5B & 13B & \textbf{4.3B}\\
\midrule
ARC-challenge & 48.55 & 61.43& 62.2 & 68.09 & 67.49 \\
Hellaswag & 71.02 & 83.32 & 82.17 & 86.62 & 85.22 \\
MMLU & 40.05 & 62.65 & 63.89 & 70.53 & 70.48 \\
TruthfulQA & 34.38 & 44.06 & 46.64 & 48.59 & 56.64 \\
WinoGrande & 66.06 & 79.24 & 76.16 & 83.35 & 82.24 \\
GSM8k & 18.72 & 40.17 & 50.84 & 58.91 & 68.50 \\
\midrule
CommonSense & 63.4 & 75.8 & 76.2 & 78.07 & 78.52 \\
\midrule
OpenLLM Leaderboard Avg. & 46.46 & 61.57 & 63.65 & 69.34 & 71.76 \\
\bottomrule
\end{tabular}
\caption{Downstream benchmarks results from four different models.}
\label{tab:benchs}
\end{table}

Table~\ref{tab:benchs} shows the results from different models. TurboSparse-Mistral-7B outperforms Gemma-2B by far, while only activating 3B parameters. TurboSparse-Mixtral-47B outperforms the original Mixtral-47B with only 4.5B parameters activated. The results demonstrate that LLMs with ReLU based intrinsic activation sparsity can keep the same or better performance while hold the significant FLOPs reduction.

\subsection{Sparsity of Sparsified Models}
\begin{figure}[!ht]
\centering
\includegraphics[width=0.6\linewidth]{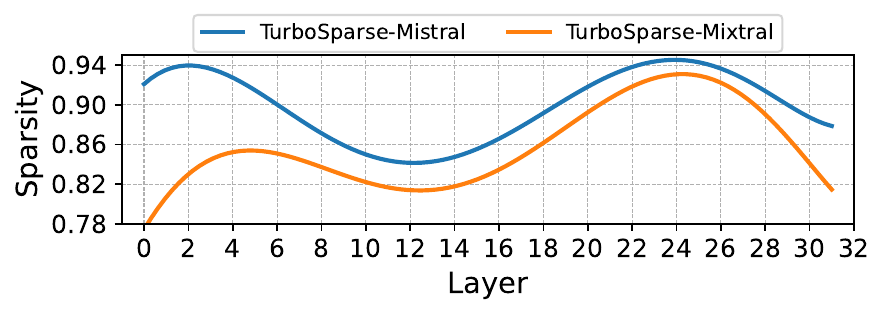}
\caption{
Sparsity of TurboSparse-Mistral-7B and TurboSparse-Mixtral-47B of different layers.
}
\label{fig:sparsity}
\end{figure}

In this subsection, we report our models' sparsity. We first profile the proportion of zero-valued activations for every layer with a general dataset(fineweb), as shown in Figure \ref{fig:sparsity}. By considering activations with a value of zero, we find that for TurboSparse-Mistral-7B, on average, has 90\% of the neurons inactive in each layer. For TurboSparse-Mixtral-47B, this percentage is slightly lower at 85\% on average for each expert FFN. Originally, Mixtral-47B would activate 2 out of 8 experts in each layer, introducing 75\% sparsity, meaning only 25\% of FLOPs needed to be computed. Furthermore, after ReLUfication, each expert will only activate 15\% of neurons. Combining these, in inference, only 3\% of parameters in each MoE layer will be activated.






\section{Practical Inference Speedup Evaluation}
In this section, we evaluate the practical acceleration in model generation achieved. During the SFT phase, we incorporate a predictor module for each FFN block. Notably, for the TurboSparse-Mixtral-47B, we train predictors for each expert. When an expert is routed, the neuron-level predictor identifies which neurons will be activated, enabling neuron-level sparse computation.

We integrate our two models with PowerInfer, which is a state-of-the-art sparsely-activated framework to evaluate the actual generation speed.
\subsection{Experiments Setting}
\paragraph{Baselines.} We take llama.cpp~\cite{llama.cpp} as our baselines for comparison. llama.cpp is the most representative inference framework. 
\paragraph{Models.} For PowerInfer and PowerInfer-2~\cite{xue2024powerinfer2}, we deployed our sparsified models, while for llama.cpp, we employed the original models for speed comparison.

\paragraph{Hardware Configurations.} All experiments were conducted on three distinct configurations:
\begin{itemize}
    \item \textbf{PC-Laptop}: Intel i9-14900HX processor, 32GB host memory (67.2 GB/s bandwidth),
    an NVIDIA RTX 4090 GPU (16GB), and PCIe 4.0 interface (64GB/s bandwidth).
    \item \textbf{PC-2080Ti}: Intel i7-12700K processor (eight 4.9GHz cores), 64GB host memory (38.4 GB/s bandwidth),
    an NVIDIA RTX 2080Ti GPU (11GB), and PCIe 3.0 interface (32GB/s bandwidth).
    \item  \textbf{OnePlus-12:} Equipped with a Snapdragon 8 Gen 3 SoC, 24 GB DRAM, and UFS 4.0 storage.
\end{itemize}
\subsection{Pure CPU Inference}
In this subsection, we focus on utilizing only the CPU for inference in our models. Due to limitations in DRAM, our evaluations are constrained to CPU performance. Table \ref{tab:cpu} presents the decoding speed results achieved with CPU-only processing for different models and settings.

The table provides a comparison of decoding speeds (in tokens per second) for various models under different settings using CPU-only inference. Overall, our ReLUfied models can achieve 2.08-2.28$\times$ speedup over the original model.
\begin{table}[t]
\centering
\caption{Decoding Speed with CPU only (tokens/s)}
\begin{tabular}{lllll}
\toprule
\textbf{Setting} & \textbf{Model} & \textbf{PowerInfer} & \textbf{llama.cpp}& \textbf{Speedup} \\ \midrule
PC-2080Ti & Mistral-7B-FP16 &  9.94 & 4.78 & 2.08$\times$\\ 
PC-2080Ti & Mixtral-47B-INT4 &  11.98 & 4.26 & 2.81$\times$\\ 
PC-Laptop & Mistral-7B-FP16  & 8.71 & 4.13 & 2.11$\times$\\
PC-Laptop & Mixtral-47B-INT4  & 16.1 & 6.91 & 2.32$\times$\\ \bottomrule
\end{tabular}
\label{tab:cpu}
\end{table}
\subsection{Hybrid GPU-CPU Inference}
In this subsection, we shift our focus to evaluating our models in a hybrid GPU-CPU computing environment, considering that most PCs are equipped with consumer-grade GPUs. Table \ref{tab:hybrid} presents the decoding speed results achieved with hybrid GPU-CPU computing for different models and settings.

The table below provides a comparison of decoding speeds (in tokens per second) for various models under different settings using a combination of GPU and CPU for inference. Overall, our models demonstrate significant speedups ranging from 2.52 to 4.64$\times$ compared to the baseline llama.cpp.
\begin{table}[t]
\centering
\caption{Decoding Speed with CPU/GPU hybrid computing (tokens/s)}
\label{tab:performance_comparison}
\begin{tabular}{lllll}
\toprule
\textbf{Setting} & \textbf{Model} & \textbf{PowerInfer} & \textbf{llama.cpp}& \textbf{Speedup} \\ \midrule
PC-2080Ti & Mistral-7B-FP16 &  35.5 & 7.64 & 4.64$\times$\\ 
PC-2080Ti & Mixtral-47B-INT4 &  22.24 & 6.63 & 3.35$\times$\\ 
PC-Laptop & Mixtral-47B-INT4  & 33.12 & 13.1 & 2.52$\times$\\ \bottomrule
\end{tabular}
\label{tab:hybrid}
\end{table}
\subsection{Deploy LLMs on mobile phones}
We also serve TurboSparse-Mixtral-47B by using PowerInfer-2 that supports LLM inference on mobile phones.
PowerInfer-2 leverages the sparse activation feature during LLM inference and introduces a computational engine on heterogeneous XPUs.
It can perform high-speed inference even when the model parameters exceed DRAM capacity.
As shown in Table \ref{tab:phone}, PowerInfer-2 achieves a 22.2$\times$ speedup using TurboSparse-Mixtral-47B inference compared to llama.cpp with the original Mixtral-47B. This significant performance gain is primarily because PowerInfer-2 can fully exploit the extremely high sparsity that TurboSparse demonstrates during inference.

\begin{table}[t]
\centering
\caption{Decoding Speed on Mobile Phones (tokens/s)}
\label{tab:mobile}
\begin{tabular}{llccc}
\toprule
\textbf{Setting} & \textbf{Model} & \textbf{PowerInfer-2} & \textbf{llama.cpp}& \textbf{Speedup} \\ \midrule
OnePlus-12 & Mixtral-47B-INT4 &  11.1 & 0.5 & 22.2$\times$\\ 
 \bottomrule
\end{tabular}
\label{tab:phone}
\end{table}
\section{Conclusion}
We propose a novel dReLU-based sparsification method that increases model sparsity to 90\% while maintaining performance, achieving a 2-5$\times$ speedup in inference. This method significantly reduces resource requirements for deploying large models, making them more environmentally friendly and accessible. This breakthrough is expected to accelerate the development of natural language processing technologies, benefiting a wider range of users. We believe that the dReLU-based sparsification method will be crucial for efficient, high-performing, and widely accessible LLMs.

{
\bibliographystyle{plain}
\bibliography{ms}
}

\appendix

\section{Appendix / supplemental material}
\subsection{Training Details of 300M models}
\label{sc:300}
In this subsection, we will introduce the details of training the 300M model, including the model architecture, types of data used, and hyperparameters. The evaluation results of the final 300M models are shown in Table~\ref{tab:1B_average}.
\begin{table}[h]
\centering
\small
\caption{Accuracy (\%) of models on evaluation datasets with average.}
\label{tab:1B_average}
\begin{tabular}{lrrrrrrrr}
\toprule
& \multicolumn{1}{l}{ARC:E} & \multicolumn{1}{l}{ARC:C} & \multicolumn{1}{l}{PIQA} & \multicolumn{1}{l}{Winogrande} & \multicolumn{1}{l}{BoolQ} & \multicolumn{1}{l}{HellaSwag} & \multicolumn{1}{l}{LAMBADA} & \multicolumn{1}{l}{Average} \\\midrule
SwiGLU & 40.03 & 22.95 & 62.68 & 52.72 & 60.92 & 31.63 & 24.34 & 42.18 \\
DReLU & 40.07 & 22.44 & 63.82 & 52.33 & 61.50 & 31.08 & 24.35 & 42.23 \\\bottomrule
\end{tabular}
\end{table}
\subsubsection{Architecture}
We adopt a similar model architecture to Llama 2~\cite{touvron2023llama} with the following details:
\begin{table}[htbp]
    \centering
    \normalsize
    \caption{ 
      Details of model architecture.
    }
    \scalebox{1.0}{
    \begin{tabular}{lcccccccc}
      \toprule
      \textbf{Hidden size}  & \textbf{Context Len}& \textbf{Heads} & \textbf{Layers} & \textbf{Vocab size}\\
  
      \midrule    
      1,024       & 2,048       & 16    & 24  & 32,000  \\
      \bottomrule
    \end{tabular}
    }
    \label{tab:model}
\end{table}

\paragraph{Activation Function and Intermediate Hidden Size.} 
We focus on dReLU and SwiGLU~\cite{shazeer2020glu} activation functions. 

\paragraph{Multi-Head Attention.} 
For Attention block, we adopt the Llama-2-7B's architecture, apply pre-normalization using RMSNorm~\cite{zhang-sennrich-neurips19} and RoPE~\cite{su2024roformer} for Positional embedding.
\subsubsection{Training Hyperparameters}
We utilize LLaMA-Factory as our training framework~\cite{zheng2024llamafactory}.
Our models are trained using AdamW optimizer~\cite{loshchilov2017decoupled}, with the following hyper-parameters: $\beta_1 = 0.9, \beta_2 = 0.95$.
We adopt a cosine learning rate schedule and we set weight decay to 0.01 and gradient clipping hyper-parameters. (see Table ~\ref{tab:hyperparameters} for more details).
\begin{table}[htbp]
    \centering
    \normalsize
    \caption{ 
      Details of optimization hyper-parameters.
    }
    \scalebox{1.0}{
    \begin{tabular}{lcccccccc}
      \toprule
      \textbf{Sequence Length}  & \textbf{Batch Size}& \textbf{Learning Rate} & \textbf{Warmup Steps} & \textbf{Hardware}\\
  
      \midrule    
      2,048       & 128       & 1e$^{-4}$    & 2000  & 4 A100-80G GPUs  \\
      \bottomrule
    \end{tabular}
    }
    \label{tab:hyperparameters}
\end{table}
\subsection{Activation Distribution Analysis of MoE Models}
\label{pe:MoE}
\begin{figure}[!ht]
\centering
\includegraphics[width=1\linewidth]{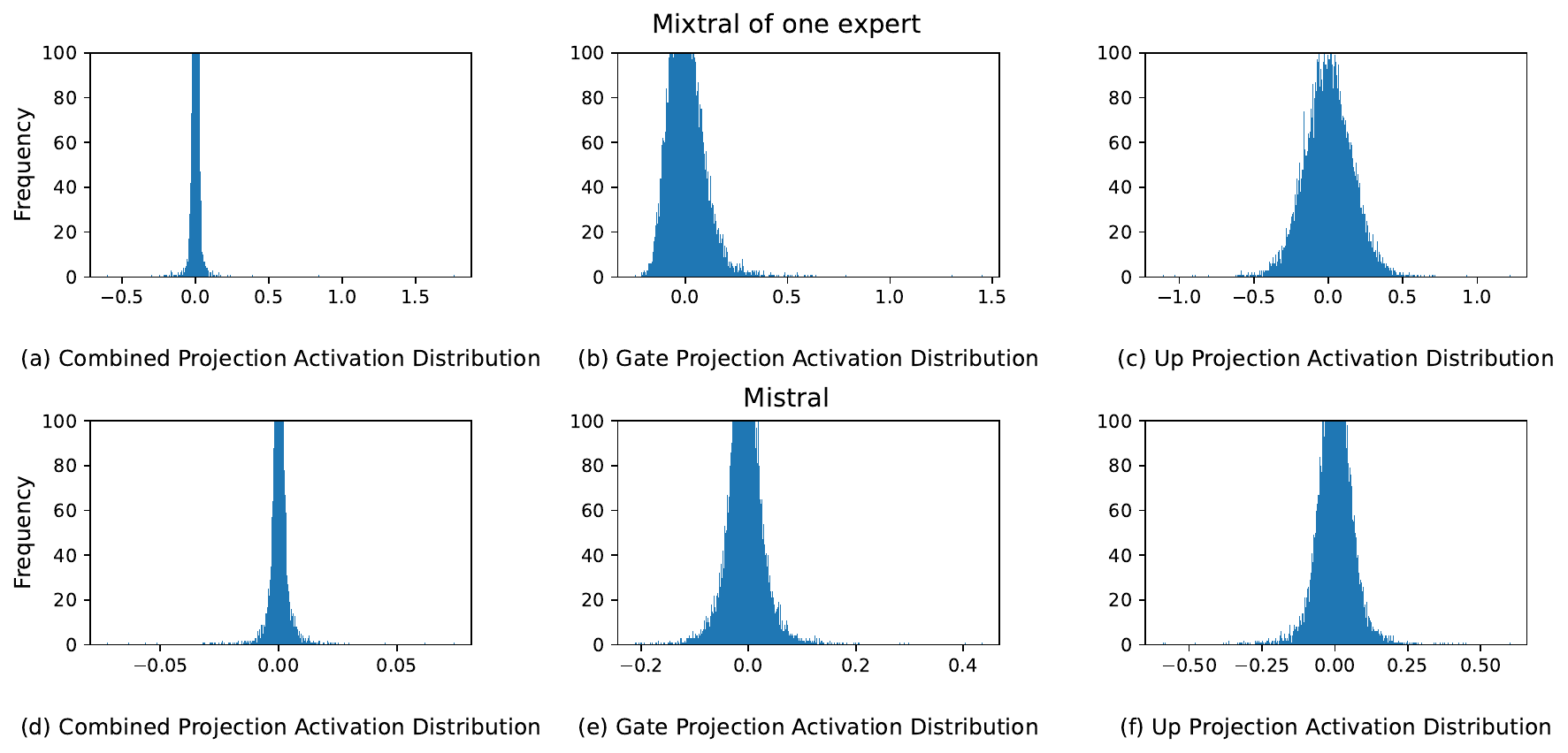}
\caption{
Post-activation distribution of Mixtral and Mistral.
}
\label{fig:activation_moe}
\end{figure}

Figure \ref{fig:activation_moe} shows the activation distribution of Mistral and Mixtral. We can see that the FFN in MoE models show the similar activation distribution compared to dense Mistral models.

\subsection{Details Performance of ReLUfied Models}
In this subsection, we present the detailed performance metrics of our ReLUfied models across various commonsense benchmarks, as shown in Table \ref{tab:bench_common}.

\begin{table}[ht!]
\centering
\begin{tabular}{ccccccc}
\toprule
& Gemma & Mistral & TurboSparse & Mixtral & TurboSparse  \\
& -2B & -7B & -Mistral-7B &  & -Mixtral-47B \\
\midrule
\# Total Params & 2B & 7B & 7B & 47B & 47B\\
\# Activate Params & 2B & 7B & 2.5B & 13B & \textbf{4.3B}\\
\midrule
SciQ & 93.8 & 96.4 & 96.4 & 96.7 & 97.9 \\
PIQA & 76.71 & 80.79 & 80.58 & 82.43 & 82.15 \\
OpenBookQA & 39.8 & 46 & 47 & 49.4 & 49 \\
ARC-Easy & 74.12 & 80.3 & 81.06 & 83.75 & 85.06 \\
\midrule
All Avg. & 71.11 & 75.87 & 76.26 & 78.07 & 78.53 \\
\bottomrule
\end{tabular}
\caption{Common benchmarks results from four different models.}
\label{tab:bench_common}
\end{table}
\section{Limitation}
Our models have only undergone continued training on 150B tokens. Compared to the 15T tokens used in pre-training for Llama-3~\cite{touvron2023llama}, the limited number of training tokens still results in some deficiencies in the model's capabilities. We are optimistic that further training can help to mitigate these shortcomings.

\section{Broader Impact}
The paper introduces a dReLU-based sparsification method and verifies its effectiveness on both dense and MoE LLMs. This approach significantly reduces computational demands, addresses environmental concerns through lower energy consumption, and helps democratize access to advanced AI technologies. We believe that our work can better support smaller organizations, educational institutions, and researchers, who previously faced barriers due to resource limitations, in accessing LLMs more easily.


\end{document}